\documentclass[nohyperref]{article}

\usepackage{microtype}
\usepackage{graphicx}
\usepackage{subfigure}
\usepackage{booktabs} 
\usepackage{array}
\usepackage{multirow}
\usepackage{colortbl}

\usepackage{hyperref}

\usepackage[accepted]{icml2022}

\usepackage{amsmath}
\usepackage{amssymb}
\usepackage{mathtools}
\usepackage{amsthm}

\usepackage[capitalize,noabbrev]{cleveref}

\theoremstyle{plain}

\theoremstyle{definition}

\theoremstyle{remark}

\usepackage[textsize=tiny]{todonotes}

\icmltitlerunning{Multi-Task Retrieval-Augmented Text Generation with Relevance Sampling}

\begin{document}

\twocolumn[
\icmltitle{Multi-Task Retrieval-Augmented Text Generation with Relevance Sampling}



\begin{icmlauthorlist}
\icmlauthor{Sebastian Hofstätter}{tu}
\icmlauthor{Jiecao Chen}{google}
\icmlauthor{Karthik Raman}{google}
\icmlauthor{Hamed Zamani}{amherst}
\end{icmlauthorlist}

\icmlaffiliation{tu}{TU Wien, Austria (work conducted during an internship at Google)}
\icmlaffiliation{google}{Google, USA}
\icmlaffiliation{amherst}{University of Massachusetts Amherst, USA}

\icmlcorrespondingauthor{Sebastian Hofstätter}{s.hofstaetter@tuwien.ac.at}


\vskip 0.3in
]



\printAffiliationsAndNotice{}  

\begin{abstract}
This paper studies multi-task training of retrieval-augmented generation models for knowledge-intensive tasks.  
We propose to clean the training set by utilizing a distinct property of knowledge-intensive generation: The connection of query-answer pairs to items in the knowledge base. We filter training examples via a threshold of confidence on the relevance labels, whether a pair is answerable by the knowledge base or not.
We train a single Fusion-in-Decoder (FiD) generator on seven combined tasks of the KILT benchmark. 
The experimental results suggest that our simple yet effective approach substantially improves competitive baselines on two strongly imbalanced tasks; and shows either smaller improvements or no significant regression on the remaining tasks. Furthermore, we demonstrate our multi-task training with relevance label sampling scales well with increased model capacity and achieves state-of-the-art results in five out of seven KILT tasks. 
\end{abstract}

\vspace{-0.7cm}
\section{Introduction}
\label{sec:intro}
\vspace{-0.1cm}
Retrieval augmented generation models are trained as a unit consisting of retrieval and generation modules
\cite{lewis2020rag}. The knowledge base accessed by the retriever module offers many benefits for practical use, such as maintainability through updates and domain adaptations. On the other hand, this setup brings additional complexity to the text generation tasks, as we now administer connections of a query-answer pair to relevant items in the knowledge base, for a more holistic view including retrieval performance \cite{zamani2022retrieval}. The coverage sparsity of relevance judgements of large collections and the resulting reliability issues are well studied, yet still a timely problem in the retrieval community \cite{zovel1998reliability, voorhees2001philosophy,craswell2021trec,hofstatter2021_trec}. This challenge is exacerbated when tasks are retroactively expanded \cite{kwiatkowski2019natural}, re-purposed \cite{bajaj2016ms} or adapt the collection \cite{kilt}.

We propose a simple yet effective approach for training retrieval-augmented models for knowledge-intensive tasks with noisy labels. 
We use a confidence score for query-answer pairs and items in the knowledge base. This confidence can be sourced from manually annotated, heuristic, or model generated aspects. We filter training examples via a threshold of confidence on the relevance labels, whether a pair is answerable by the knowledge base or not.
With this we aim to reduce noise in the training process, and produce better results with fewer training examples.

To study our training approach, we use a fixed T5-based dense retrieval module \cite{ni2021t5xRetrieval} and train a Fusion-in-Decoder (FiD) generator \cite{izacard2020fid} on multiple tasks of the KILT benchmark \cite{kilt}. KILT aggregates and heuristically maps many different English Wikipedia-based generation tasks to a single Wikipedia snapshot, which introduces considerable noise in the label quality, due to the time-shifted nature of the task creations. 

We apply our confidence threshold on relevance label filtering to remove a training example if no knowledge item could be identified as sufficiently relevant from the existing labels. Because of the time shifted knowledge base, if an answer is not available in the new passage text anymore, we have a lower confidence, that the query can be answered at all given the new passages. After this step, we apply downsampling on imbalanced tasks for a balanced multi-task training on all the seven tasks of KILT that have passage mappings; spanning open domain QA, slot filling, fact verification, and dialogue categories: HotpotQA \cite{yang2018hotpotqa},
TriviaQA \cite{joshi2017triviaqa},
Natural Questions (NQ) \cite{kwiatkowski2019natural}, T-REx \cite{elsahar2018trex},
Zero Shot RE (zsRE) \cite{levy2017zsre}), FEVER \cite{thorne2018fever}, and Wizard of Wikipedia (WoW) \cite{dinan2018wizard}. 

Furthermore, we demonstrate the robustness of our sampling strategy by creating an alternative to the prevalent original, aggregation method from Wikipedia paragraphs to retrievable units. Finally, we study the impact of our training method on increased capacities of the generator backbone.

We find that our training strategy significantly improves the effectiveness on the two strongly imbalanced datasets: TriviaQA (+ 12.7 EM) and T-REx (+4.9 Accuracy). This leads to a new state-of-the-art in TriviaQA, and is competitive in T-REx compared to more specialized models. It also statistically significant improves two out of the remaining five tasks, albeit at a smaller rate.
When scaling up the T5 backbone of the FiD model with our multi-task sampling technique from T5-Base to T5-Large and T5-XL, we observe expected quality gains across all our evaluated tasks and outperform the state-of-the-art on a total of five out of seven KILT tasks on the official leaderboard.

\begin{figure}[t]
    \includegraphics[clip,trim={0.6cm 0.7cm 0.2cm 0.4cm}, width=0.48\textwidth]{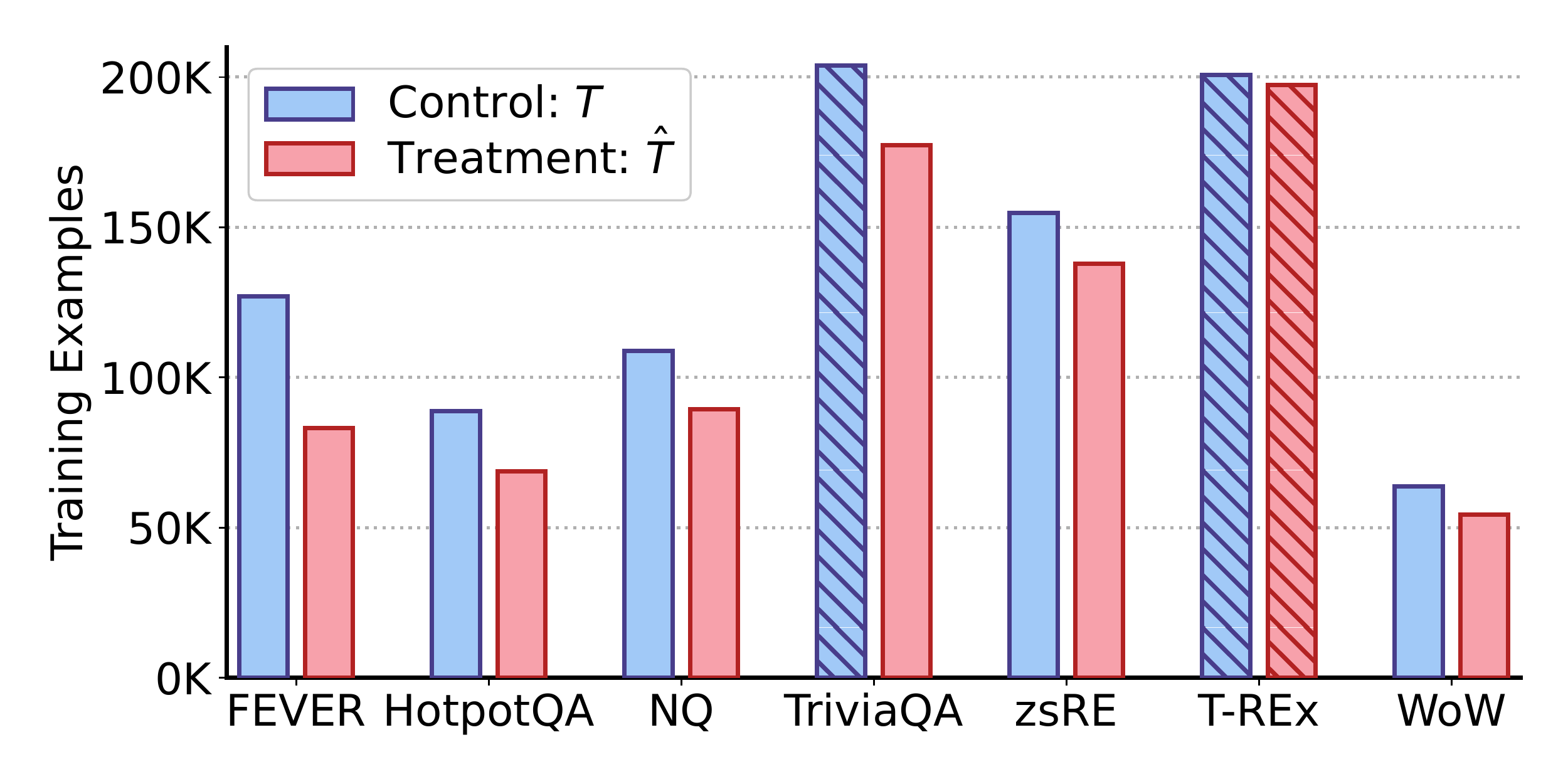}
    \centering
    \vspace{-0.5cm}
    \caption{Training examples per task and sampling method. \textit{Hatched bars indicate downsampling with potentially more training data available.}}
    \label{fig:dataset_sampling_counts}
\vspace{-0.2cm}
\end{figure}

\section{Relevance-Based Confidence Sampling}
\label{sec:method}

%
%
%

The main goal in retrieval-augmented generation is to generate an answer string $a$ given a query $q$; with a secondary goal of identifying a set of relevant passages $P$ from a collection $C$, which are the source of the answer. In a dataset, the relevant passage set $P^{(q,a)}_t$ using a threshold $t$, is:
\begin{equation}
\begin{aligned} 
P^{(q,a)}_t = \big\{p \ \big\vert \ \Phi(p, q, a) > t, \forall \ p \in C \big\}
\end{aligned}
\end{equation}
where $\Phi$ is a mapping function between a passage, query and answer triple, returning a confidence value, whether this passage is relevant or not. Only if the confidence is higher than our set threshold $t$, do we include the passage in the set. From the view of a dataset creator, it is usually unfeasible to conduct annotations for all possible pairs, therefore, those pairings without annotations return a null confidence for relatedness, even if it might be related.

As dataset creation is a very costly operation, many works adapt and evolve existing datasets. When knowledge intensive datasets are evolved, the query-answer pair may stay the same while the confidence values of the passage connections change. 
Therefore, we hypothesise it is beneficial not to include all possible training examples, rather only take into account query-answer pairs, where a higher confidence threshold on the relevance label is set, to reduce noise. A low or no confidence value might indicate a low quality query answer pair. 

Starting from the training set ${T}$, which includes all possible pairs $(q, a)$, we define a filtered version $\Hat{{T}_t}$ as follows: 
\begin{equation}
\begin{aligned} 
\Hat{{T}_t} = \big\{(q, a) \ \big\vert \ \exists  \ p \in P^{(q,a)}_t, \forall  \ (q, a) \in {T} \big\}
\end{aligned}
\end{equation}
where we need to define a threshold $t$ as our sampling boundary. The boundary needs to be adapted to the properties of the applied task.

In this work we apply our sampling approach on KILT which conducted a heuristic mapping process of passages for given query-answer pairs. We implement the confidence mapping $\Phi$ as the BLEU score in their mapping and set the threshold $t$ to be $>0$, filtering all pairs, where no overlap in the previously annotated document was found. Our sampling is not limited to KILT and could be extended to other resources with a similar setup or by mining weakly-supervised relevance signals, as proposed by \citet{asai2021evidentiality}, and filtering for example based on the confidence of the labelling model.

\section{Experiment Design}

%
%

\begin{figure}[t]
    \includegraphics[clip,trim={0.6cm 0.7cm 0.2cm 0.4cm}, width=0.48\textwidth]{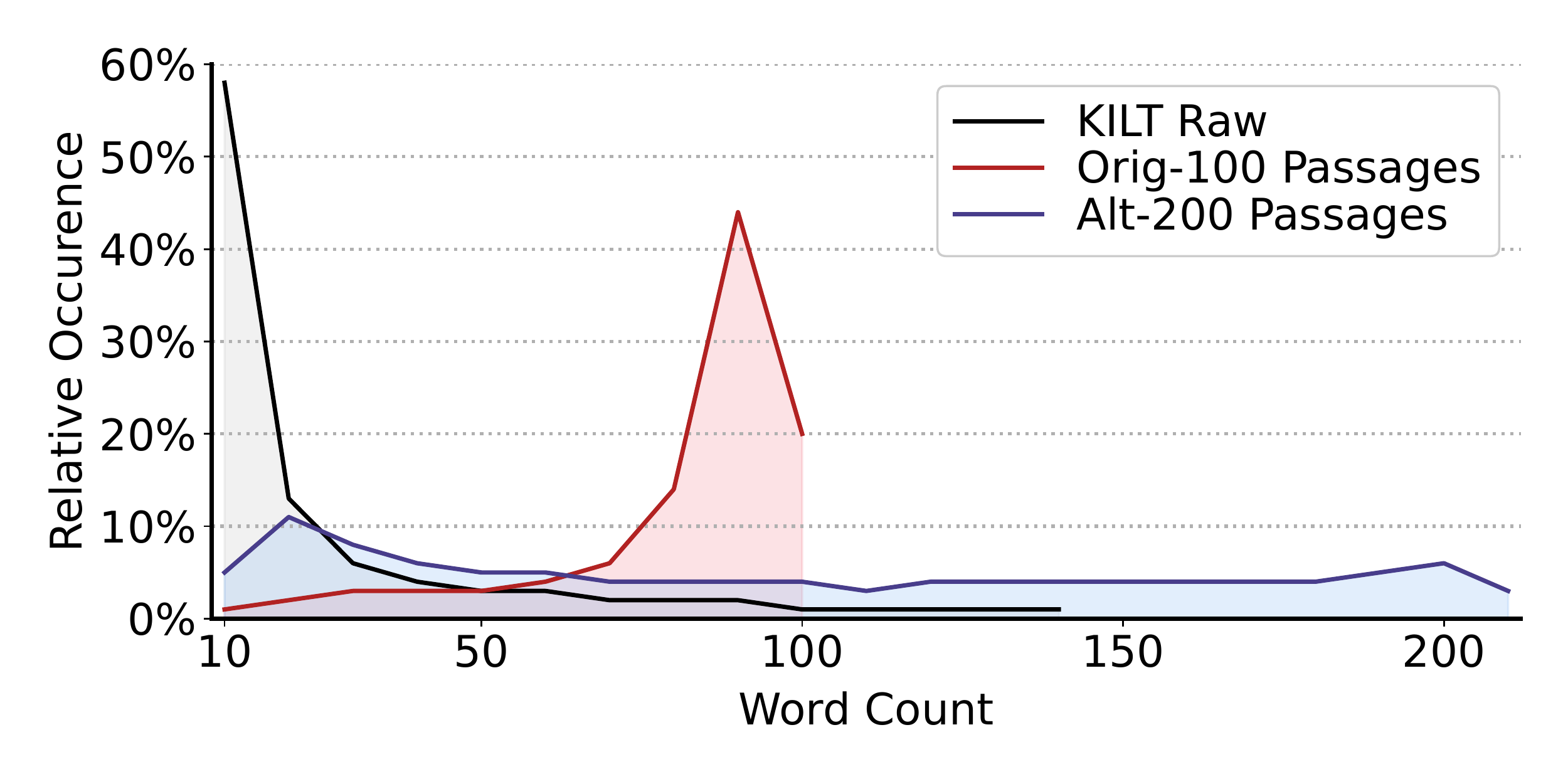}
    \centering
    \vspace{-0.6cm}
    \caption{Statistics of the passage lengths of the raw KILT texts, its original chunking (Orig-100) and our alternative approach (Alt-200). \textit{The word counts are binned to 10 words.}}
    \label{fig:retrievable_unit_lengths}
    \vspace{-0.4cm}
\end{figure}

\paragraph{KILT multi-task training.} 
We train a single generator model on multiple tasks of the KILT benchmark, most of which already provide training sets of similar magnitude (50 to 150 thousand), except for: TriviaQA (1.8 million) and T-REx (12.5 million), accounting for 96\% of training examples.
FiD training requires considerable hardware resources, therefore we decided to downsample the oversized datasets, rather than upsample the others.\footnote{A T5-Base training run until convergence requires roughly one TPU month of compute with our downsample strategies.}

Figure \ref{fig:dataset_sampling_counts} shows the number of query-answer pairs available per sampling method: our control $T$ and treatment $\hat{T}$. We apply our filtering before downsampling oversized tasks to balance our multi-task training set. For our unified training we downsample oversized tasks to 200K examples, then combine and shuffle all tasks. 
By applying the relevance-label filter method $\hat{T}$ we reduce our training set size by 25\% (or 140K fewer examples) to 808K examples compared to our control.

\begin{table*}[t]
\vspace{-0.3cm}
    \caption{Comparing sampling strategies and model capacity scaling for multi-task training on the KILT dev set. Highest result in bold. Our results are averaged over the last 10 checkpoints with a 95\% confidence interval shown in gray.}
    \vspace{0.1cm}
    \label{tab:sampling_ablation}
    \setlength\tabcolsep{2.7pt}
    \begin{tabular}{lll!{\color{gray}\vrule}lll!{\color{gray}\vrule}l!{\color{gray}\vrule}ll!{\color{gray}\vrule}l}
           \toprule

  & \multirow{3}{*}{\textbf{Model}} & \multirow{3}{*}{\textbf{LM}} & 
  \multicolumn{3}{c!{\color{lightgray}\vrule}}{\textbf{Open Domain QA}} &
  \textbf{Fact} &
  \multicolumn{2}{c!{\color{lightgray}\vrule}}{\textbf{Slot Filling}} &
  \textbf{Dialog} \\
  & & & 
  \multicolumn{3}{c!{\color{lightgray}\vrule}}{\textit{EM}} &
  \textit{Accuracy} &
  \multicolumn{2}{c!{\color{lightgray}\vrule}}{\textit{Accuracy}} &
   \textit{F1} \\  
  & & & NQ & HotpotQA & TriviaQA & FEVER & T-REx & zsRE & WOW \\
 \midrule

\multicolumn{4}{l}{\textbf{Related Methods}} \\
\textcolor{gray}{1} & RAG \cite{kilt} & BART-L & 44.4 & 27.0 & 71.3 & 86.3 & 59.2 & 44.7 & 13.1 \\
\textcolor{gray}{2} & DPR+FiD \cite{piktus2021oyster} & T5-Base & 55.0 & {38.0} & 71.4 & 90.9 & 80.9 & {72.4} & 16.1 \\
\textcolor{gray}{3} & KGI \cite{glass2021robust} & BART-L & -- & -- & --& -- & 84.0 & 71.3 & -- \\
\textcolor{gray}{4} & Re2G \cite{anon2022re2g} & BART-L & 46.7 & -- & 74.0 & {91.1} & \textbf{86.6} & -- & 19.4 \\
\midrule
\multicolumn{4}{l}{\textbf{Ours (DPR-100 passages)}} \\
\textcolor{gray}{5} & GTR + FiD with control $T$ & T5-Base        & 54.1 {\small\color{darkgray}$\pm$.3 } & 31.1 {\small\color{darkgray}$\pm$.2 } & 65.0 {\small\color{darkgray}$\pm$.6 } & 89.8 {\small\color{darkgray}$\pm$.2 } & 78.0 {\small\color{darkgray}$\pm$.5 } & 70.7 {\small\color{darkgray}$\pm$.4 } & 19.8 {\small\color{darkgray}$\pm$.2 } \\
\textcolor{gray}{6} & GTR + FiD with treatment $\hat{T}$ & T5-Base & 54.4 {\small\color{darkgray}$\pm$.3 } & 31.0 {\small\color{darkgray}$\pm$.2 } & 78.1 {\small\color{darkgray}$\pm$.2 } & 89.6 {\small\color{darkgray}$\pm$.4 } & 82.9 {\small\color{darkgray}$\pm$.1 } & 71.6 {\small\color{darkgray}$\pm$.3 } & 19.5 {\small\color{darkgray}$\pm$.2 } \\
\midrule
\arrayrulecolor{lightgray}
\multicolumn{4}{l}{\textbf{Ours (Alt-200 passages)}} \\
\textcolor{gray}{7} & GTR + FiD with control $T$ & T5-Base        & 55.1 {\small\color{darkgray}$\pm$.3 } & 31.6 {\small\color{darkgray}$\pm$.2 } & 65.7 {\small\color{darkgray}$\pm$.6 } & 89.8 {\small\color{darkgray}$\pm$.4 } & 77.6 {\small\color{darkgray}$\pm$.3 } & 70.2 {\small\color{darkgray}$\pm$.2 } & 20.1 {\small\color{darkgray}$\pm$.2 } \\
\midrule
\textcolor{gray}{8} & \multirow{3}{*}{GTR + FiD with treatment $\hat{T}$} & T5-Base & 56.0 {\small\color{darkgray}$\pm$.3 } & 31.8 {\small\color{darkgray}$\pm$.2 } & 78.4 {\small\color{darkgray}$\pm$.2 } & 89.6 {\small\color{darkgray}$\pm$.4 } & 82.5 {\small\color{darkgray}$\pm$.1 } & 71.4 {\small\color{darkgray}$\pm$.3 } & 19.9 {\small\color{darkgray}$\pm$.2 } \\
\textcolor{gray}{9} &   & T5-Large                                        & 60.7 {\small\color{darkgray}$\pm$.5 } & 35.4 {\small\color{darkgray}$\pm$.2 } & 81.8 {\small\color{darkgray}$\pm$.2 } & 92.1 {\small\color{darkgray}$\pm$.2 } & 82.9 {\small\color{darkgray}$\pm$.1 } & 72.9 {\small\color{darkgray}$\pm$.4 } & 19.9 {\small\color{darkgray}$\pm$.2 } \\
\textcolor{gray}{10} &   & T5-XL & \textbf{62.9} {\small\color{darkgray}$\pm$.4 } & \textbf{39.0} {\small\color{darkgray}$\pm$.2 } & \textbf{84.3} {\small\color{darkgray}$\pm$.2 } & \textbf{92.8} {\small\color{darkgray}$\pm$.4 } & 84.1 {\small\color{darkgray}$\pm$.2 } & \textbf{75.2} {\small\color{darkgray}$\pm$.3 } & \textbf{21.0} {\small\color{darkgray}$\pm$.3 } \\
\arrayrulecolor{black}
\bottomrule
\end{tabular}
    \vspace{-0.3cm}
\end{table*}

\paragraph{Alternative retrievable units.}

%
%

The initial KILT release offers a fine-granular view on the Wikipedia collection. The raw paragraph collection contains 111.4 million items, with a strong concentration of very short sequences ($< 20$ words), as shown in Figure \ref{fig:retrievable_unit_lengths}. This is not a practical number of passages to index with dense retrieval methods (as the memory requirement is determined by the number of passages). A standard aggregation approach, proposed by \citet{karpukhin2020dense} for DPR is to aggregate the raw data to passages of up to 100 words. It adds as many words to a new passage until 100 words or a changed document are reached, which breaks most paragraphs at the boundaries in half. Therefore, we propose an alternative aggregation strategy to relax the strict length requirement and favoring not to break up paragraphs. We aggregate whole raw-paragraphs until they reach 200 words, or start a new passage if they do not fit. This change results in a very different length distribution, as shown in Figure \ref{fig:retrievable_unit_lengths}. It results in only 27.7 million passages, compared to 35.7 million of the original chunking. In both cases the title of the page is added to all passages.

\vspace{-0.3cm}
\paragraph{Implementation.}

All our experiments are based on the T5X framework \cite{roberts2022t5x}. We use a fixed GTR-Base dense retrieval model \cite{ni2021t5xRetrieval}, which is pre-trained on the MSMARCO passage retrieval task \cite{bajaj2016ms} and has been shown to generalize well on the BEIR benchmark \cite{thakur2021beir}. We train an FiD model \cite{izacard2020fid} using T5 v1.1 as language model backbone \cite{raffel2020exploring} on TPUs. We attach task specific markers to the input for the multi-task training. We cap the input at 384 tokens (combined query and passage) and a maximum of 64 output tokens. For training we use a batch size of 128 with 50 retrieved passages, and a learning rate of 10$^{-3}$ with the Adafactor optimizer \cite{shazeer2018adafactor}. We do not tune our models to a specific checkpoint, rather train them all for 50K steps. The only special case is T5-XL, which uses a learning rate of $5*10^{-4}$ and is trained for 30K steps. We use beam search with a beam size of 4 for the decoding. 

\vspace{-0.3cm}
\paragraph{Evaluation.} To reduce the noise in our results, we present the mean and a 95\% confidence interval measured with a t-statistic of the last 10 checkpoints (every thousand steps from 40K to 50K training steps; 20K to 30K for T5-XL).

\begin{table*}[t]
\vspace{-0.3cm}
    \caption{Comparing our models with related work on the KILT test set via the leaderboard (as of July 3$^\text{rd}$ 2022). Highest result in bold.}
    \vspace{0.1cm}
    \label{tab:leaderboard}
    \setlength\tabcolsep{3.5pt}
    \begin{tabular}{lll!{\color{gray}\vrule}ccc!{\color{gray}\vrule}c!{\color{gray}\vrule}cc!{\color{gray}\vrule}c}
           \toprule

  & \multirow{3}{*}{\textbf{Model}} & \multirow{3}{*}{\textbf{Generator}} & 
  \multicolumn{3}{c!{\color{lightgray}\vrule}}{\textbf{Open Domain QA}} &
  \textbf{Fact} &
  \multicolumn{2}{c!{\color{lightgray}\vrule}}{\textbf{Slot Filling}} &
  \textbf{Dialog} \\
  & & & 
  \multicolumn{3}{c!{\color{lightgray}\vrule}}{\textit{EM}} &
  \textit{Acc.} &
  \multicolumn{2}{c!{\color{lightgray}\vrule}}{\textit{Accuracy}} &
   \textit{F1} \\  
  & & & NQ & HotpotQA & TriviaQA & FEVER & T-REx & zsRE & WOW \\
 \midrule

\multicolumn{4}{l}{\textbf{Top Leaderboard Entries}} \\
\textcolor{gray}{1} & RAG \cite{kilt} & BART-Large & 44.4 & 27.0 & 71.3 & 86.3 & 59.2 & 44.7 & 13.1 \\
\textcolor{gray}{2} & DPR + FiD \cite{piktus2021oyster} & T5-Base & 51.6 & 38.3 & 72.7 & 89.0 & 82.2 & 74.0 & 15.7 \\
\textcolor{gray}{3} & KGI \cite{glass2021robust} & BART-Large & 45.2 & -- & 61.0 & 85.6 & 84.4 & 72.6 & 18.6 \\
\textcolor{gray}{4} & Re2G \cite{anon2022re2g} & BART-Large & 51.7 & -- & 76.3 & 89.6 & \textbf{87.7} & -- & 18.9 \\
\textcolor{gray}{5} & Hindsight \cite{paranjape2021hindsight}  & BART-Large & -- & -- & -- & -- & -- & -- & 19.2 \\
\textcolor{gray}{6} & SEAL+FiD \cite{bevilacqua2022autoregressive} & T5-? & 53.7 & \textbf{40.5} & 70.9 & 89.5 & 83.7 & 74.7 & 18.3 \\

\midrule
\multicolumn{4}{l}{\textbf{Ours (Alt-200 passages)}} \\
\textcolor{gray}{7} & \multirow{2}{*}{GTR + FiD with treatment $\hat{T}$} & T5-Base & 52.4 & 30.1 & 78.9 & 87.1 & 83.4 & 81.5 & 18.4 \\
\textcolor{gray}{8} &   & T5-XL & \textbf{61.2} & 39.1 & \textbf{84.6} & \textbf{92.3} & 85.2 & \textbf{83.7} & \textbf{20.6} \\
\arrayrulecolor{black}
\bottomrule
\end{tabular}
    \vspace{-0.3cm}
\end{table*}

\vspace{-0.2cm}
\section{Results}
\label{sec:results}
\vspace{-0.1cm}

In this section we present and discuss our experimental results. An important note with every use of the KILT benchmark is that the numbers presented here are only comparable to other works also based on the KILT benchmark and not the original versions of the individual tasks. This is due to the changed collection as well as changed query sets, as described by \citet{kilt}.

The results on the KILT dev set are shown in Table \ref{tab:sampling_ablation}. In the first section (lines 1-4) we show related works, which also report KILT-based scores: RAG \cite{lewis2020rag}, as evaluated by \citet{kilt};
DPR + FiD \cite{piktus2021oyster};
KGI \cite{glass2021robust}; and
Re2G \cite{anon2022re2g}. We present our results using the original passage units in the second (lines 5 \& 6) and our alternative retrieval units in the third section (lines 7 \& 8). In both sections we compare the random downsampling and our proposed relevance-label guided sampling strategy. 

\vspace{-0.3cm}
\paragraph{Sampling strategies.}

First, we focus on the two strongly imbalanced tasks (TriviaQA and T-REx), which had their training examples change the most under our relevance-label sampling strategy: We see that for both passage variants, both tasks improve considerably with the proposed sampling. Comparing lines 7 \& 8 we observe a gain for TriviaQA of 12.7 EM and 4.9 Accuracy for T-REx.
For the other tasks on our alternative passage-units, we observe small, but significant gains on NQ, and zsRE. The other tasks FEVER, WOW, and HotpotQA only result in non-significant changes inside the 95\% confidence interval. 

\vspace{-0.15cm}
\paragraph{Retrievable units.}
To observe the impact of our alternative passage aggregation strategy we need to compare the pairs of lines 5 \& 7 as well as lines 6 \& 8. Even though the properties of the two passage collections are very different, the results of the retrieval augmented generation are very similar. Our alternative approach is slightly better on NQ, HotpotQA, and WOW. Overall, we also notice, that our passage sampling strategy works slightly better on the alternative passages, resulting in the best overall results of our ablation. Therefore, we select this combination (line 8) for the following experiments.

\vspace{-0.15cm}
\paragraph{Scaling the generator capacity.}

In most NLP settings, increasing the capacity of a pre-trained model leads to effectiveness gains, at the cost of efficiency. Given how the related methods use varying generator capacities (such as BART-Large \cite{lewis2019bart}), we want to understand and measure the implications of scaling up the generator for our alternative passage aggregation and relevance-label sampling strategy $\hat{T}$. We show these results in Table \ref{tab:sampling_ablation} for T5-Base (line 8), T5-Large (line 9), and T5-XL (line 10).

We find that scaling the model size and compute resource consistently improve results over all tasks. This is an expected result. Nevertheless, we wanted to confirm that our sampling improvement is not just beneficial in a smaller setting.

\paragraph{Leaderboard comparison.} We submitted a T5-Base and T5-XL version of the FiD model with our relevance sampling to the official KILT leaderboard\footnote{The leaderboard is available at: \\ \url{https://eval.ai/web/challenges/challenge-page/689}} for a blind evaluation and present the results in Table \ref{tab:leaderboard}. 
Compared to related methods our FiD model with T5-Base is already state of the art on two tasks (TriviaQA and zsRE). Our T5-XL version sets a new state of the art ceiling on a total of five KILT tasks. We outperform the previous best methods by: NQ +7.4 EM, 
TriviaQA +6.4 EM,
FEVER +2.7 Accuracy, 
ZS-RE +9 Accuracy, 
and WoW +0.05 F1. 
We only come in second place on HotpotQA (-1.4 EM) and T-REx (-2.5 Accuracy). This might be attributable to our handicapped zero shot retriever, as HotpotQA is challenging for retrieval models; and downsampling of the T-REx training data, as the related methods are trained exclusively on the single task, without the need for training data adjustments. 

Overall, these results are a strong indicator for the viability and usefulness of our relevance-label sampling strategy considering that it has access to 140K fewer training examples than the baseline. 
We want to emphasize that when we compare our already competitive results to related work our approach is handicapped in a few key areas: \textbf{1)} we are not training the retriever (which is out of scope, but orthogonal to our work and should lead to further improvements); \textbf{2)} we are training a single model, which gives us less chance to overfit on a single task; \textbf{3)} we do not employ multiple training loops, index updates, or knowledge distillation. Therefore, we conclude that multi-task training is a viable option for the community to build upon going forward. 

\vspace{-0.1cm}
\paragraph{Are we just gaming the benchmark?} A valid concern we need to raise is whether we are really improving the quality of the model, or simply moving the training set construction closer to the way the tests sets have been constructed by \citet{kilt}. The KILT test sets filter an average 18\% of queries compared to their original task versions. \citet{kilt} removed a query if not at least one of the answers could be mapped to a passage at least once. Crucially, if one of the answers is partially mappable, all the other answers for this query were also kept as valid. Our analysis shows that, while the average ratio of mapped answers increases compared to the raw training data, especially exact mapped answers still only account for 10\% to 67\% of available answers. Therefore, we argue that we are not gaming the benchmark, as we exclusively select mapped query-answer pairs for our training, which differs from the test set construction. For a conclusive answer to this question future work should evaluate our training procedure on other, independently created, evaluation tasks. A setup which is increasingly common in the neural retrieval community \cite{ni2021t5xRetrieval,Hofstaetter2022_colberter}. 


\section{Related Work}
\label{sec:related}


\paragraph{Multi-task training.}

To the best of our knowledge, the multi-task focus of the KILT community so far has been on the retriever module and not the answer generator. The foundational retrieval augmented architectures FiD \cite{izacard2020fid}, RAG \cite{lewis2020rag}, and REALM \cite{guu2020realm} are trained on individual tasks. In their initial baseline setup \citet{kilt} already studied the impact of multi-task retrieval training; \citet{maillard2021multi} continued to study various configurations for KILT multi-task single-model retrieval. \citet{lewis2021paq} trained the RePAQ-retriever system on multiple tasks, but for their FiD defer mechanism used task-specific FiD checkpoints.
For context-given question answering \citet{khashabi2020unifiedqa} trained UnifiedQA on multiple QA tasks. 

\paragraph{Improved RAG training.}
Many of the recent papers improving RAG-style models optimized end-to-end processes (f.e. EMDR2 \cite{singh2021emdr}), ensembling multiple modules (f.e. R2-D2 \cite{fajcik2021r2d2}), or creating multiple training loops to update the indexed documents multiple times (f.e.
Hindsight \cite{paranjape2021hindsight}).
Our approach differs, as we focus on the selection of the available training data in a multi-task setting. For more information on retrieval-enhanced machine learning models, we refer the reader to \citet{zamani2022retrieval}.

\section{Conclusion}
\label{sec:conclusion}

We proposed a simple yet effective approach for multi-task training of the FiD retrieval-augmented generation model on the KILT benchmark. We cleaned (and downsampled were necessary) the training set by removing query-answer pairs with low relevance confidence. We demonstrated that this approach substantially improves two imbalanced tasks, and has a smaller benefit on two of the remaining five tasks. By scaling the model capacity we achieve state-of-the-art results on five KILT tasks evaluated by the leaderboard.

\section*{Acknowledgements}
This research was supported in part by the Google Visiting Scholar program and in part by the Center for Intelligent Information Retrieval. Any opinions, findings, and conclusions or recommendations expressed in this material are those of the authors and do not necessarily reflect those of the sponsors. 
We would like to thank Jianmo Ni for helping us setting up T5X retrieval.

\bibliography{main}

\begin{thebibliography}{36}
\providecommand{\natexlab}[1]{#1}
\providecommand{\url}[1]{\texttt{#1}}
\expandafter\ifx\csname urlstyle\endcsname\relax
  \providecommand{\doi}[1]{doi: #1}\else
  \providecommand{\doi}{doi: \begingroup \urlstyle{rm}\Url}\fi

\bibitem[Anonymous(2022)]{anon2022re2g}
Anonymous.
\newblock Re2g: Retrieve, rerank, generate.
\newblock \emph{ACL Submission (OpenReview)}, 2022.

\bibitem[Asai et~al.(2021)Asai, Gardner, and Hajishirzi]{asai2021evidentiality}
Asai, A., Gardner, M., and Hajishirzi, H.
\newblock Evidentiality-guided generation for knowledge-intensive nlp tasks.
\newblock \emph{arXiv preprint arXiv:2112.08688}, 2021.

\bibitem[Bajaj et~al.(2016)Bajaj, Campos, Craswell, Deng, Gao, Liu, Majumder,
  McNamara, Mitra, Nguyen, et~al.]{bajaj2016ms}
Bajaj, P., Campos, D., Craswell, N., Deng, L., Gao, J., Liu, X., Majumder, R.,
  McNamara, A., Mitra, B., Nguyen, T., et~al.
\newblock Ms marco: A human generated machine reading comprehension dataset.
\newblock \emph{arXiv preprint arXiv:1611.09268}, 2016.

\bibitem[Bevilacqua et~al.(2022)Bevilacqua, Ottaviano, Lewis, Yih, Riedel, and
  Petroni]{bevilacqua2022autoregressive}
Bevilacqua, M., Ottaviano, G., Lewis, P., Yih, W.-t., Riedel, S., and Petroni,
  F.
\newblock Autoregressive search engines: Generating substrings as document
  identifiers.
\newblock \emph{arXiv preprint arXiv:2204.10628}, 2022.

\bibitem[Craswell et~al.(2021)Craswell, Mitra, Yilmaz, Campos, Voorhees, and
  Soboroff]{craswell2021trec}
Craswell, N., Mitra, B., Yilmaz, E., Campos, D., Voorhees, E.~M., and Soboroff,
  I.
\newblock Trec deep learning track: Reusable test collections in the large data
  regime.
\newblock In \emph{Proceedings of the 44th International ACM SIGIR Conference
  on Research and Development in Information Retrieval}, pp.\  2369--2375,
  2021.

\bibitem[Dinan et~al.(2018)Dinan, Roller, Shuster, Fan, Auli, and
  Weston]{dinan2018wizard}
Dinan, E., Roller, S., Shuster, K., Fan, A., Auli, M., and Weston, J.
\newblock Wizard of wikipedia: Knowledge-powered conversational agents.
\newblock \emph{arXiv preprint arXiv:1811.01241}, 2018.

\bibitem[Elsahar et~al.(2018)Elsahar, Vougiouklis, Remaci, Gravier, Hare,
  Laforest, and Simperl]{elsahar2018trex}
Elsahar, H., Vougiouklis, P., Remaci, A., Gravier, C., Hare, J., Laforest, F.,
  and Simperl, E.
\newblock T-rex: A large scale alignment of natural language with knowledge
  base triples.
\newblock In \emph{Proceedings of the Eleventh International Conference on
  Language Resources and Evaluation (LREC 2018)}, 2018.

\bibitem[Fajcik et~al.(2021)Fajcik, Docekal, Ondrej, and Smrz]{fajcik2021r2d2}
Fajcik, M., Docekal, M., Ondrej, K., and Smrz, P.
\newblock R2-d2: A modular baseline for open-domain question answering.
\newblock \emph{arXiv preprint arXiv:2109.03502}, 2021.

\bibitem[Glass et~al.(2021)Glass, Rossiello, Chowdhury, and
  Gliozzo]{glass2021robust}
Glass, M., Rossiello, G., Chowdhury, M. F.~M., and Gliozzo, A.
\newblock Robust retrieval augmented generation for zero-shot slot filling.
\newblock \emph{arXiv preprint arXiv:2108.13934}, 2021.

\bibitem[Guu et~al.(2020)Guu, Lee, Tung, Pasupat, and Chang]{guu2020realm}
Guu, K., Lee, K., Tung, Z., Pasupat, P., and Chang, M.-W.
\newblock Realm: Retrieval-augmented language model pre-training.
\newblock \emph{arXiv preprint arXiv:2002.08909}, 2020.

\bibitem[Hofst{\"a}tter et~al.(2021)Hofst{\"a}tter, Sertkan, and
  Hanbury]{hofstatter2021_trec}
Hofst{\"a}tter, S., Sertkan, M., and Hanbury, A.
\newblock Tu wien at trec dl and podcast 2021: Simple compression for dense
  retrieval.
\newblock 2021.

\bibitem[Hofst{\"a}tter et~al.(2022)Hofst{\"a}tter, Khattab, Althammer,
  Sertkan, and Hanbury]{Hofstaetter2022_colberter}
Hofst{\"a}tter, S., Khattab, O., Althammer, S., Sertkan, M., and Hanbury, A.
\newblock Introducing neural bag of whole-words with colberter: Contextualized
  late interactions using enhanced reduction.
\newblock 2022.
\newblock \doi{10.48550/ARXIV.2203.13088}.

\bibitem[Izacard \& Grave(2020)Izacard and Grave]{izacard2020fid}
Izacard, G. and Grave, E.
\newblock Leveraging passage retrieval with generative models for open domain
  question answering.
\newblock \emph{arXiv preprint arXiv:2007.01282}, 2020.

\bibitem[Joshi et~al.(2017)Joshi, Choi, Weld, and
  Zettlemoyer]{joshi2017triviaqa}
Joshi, M., Choi, E., Weld, D.~S., and Zettlemoyer, L.
\newblock Triviaqa: A large scale distantly supervised challenge dataset for
  reading comprehension.
\newblock \emph{arXiv preprint arXiv:1705.03551}, 2017.

\bibitem[Karpukhin et~al.(2020)Karpukhin, O{\u{g}}uz, Min, Lewis, Wu, Edunov,
  Chen, and Yih]{karpukhin2020dense}
Karpukhin, V., O{\u{g}}uz, B., Min, S., Lewis, P., Wu, L., Edunov, S., Chen,
  D., and Yih, W.-t.
\newblock Dense passage retrieval for open-domain question answering.
\newblock \emph{arXiv preprint arXiv:2004.04906}, 2020.

\bibitem[Khashabi et~al.(2020)Khashabi, Min, Khot, Sabharwal, Tafjord, Clark,
  and Hajishirzi]{khashabi2020unifiedqa}
Khashabi, D., Min, S., Khot, T., Sabharwal, A., Tafjord, O., Clark, P., and
  Hajishirzi, H.
\newblock Unifiedqa: Crossing format boundaries with a single qa system.
\newblock \emph{arXiv preprint arXiv:2005.00700}, 2020.

\bibitem[Kwiatkowski et~al.(2019)Kwiatkowski, Palomaki, Redfield, Collins,
  Parikh, Alberti, Epstein, Polosukhin, Devlin, Lee,
  et~al.]{kwiatkowski2019natural}
Kwiatkowski, T., Palomaki, J., Redfield, O., Collins, M., Parikh, A., Alberti,
  C., Epstein, D., Polosukhin, I., Devlin, J., Lee, K., et~al.
\newblock Natural questions: a benchmark for question answering research.
\newblock \emph{Transactions of the Association for Computational Linguistics},
  7:\penalty0 453--466, 2019.

\bibitem[Levy et~al.(2017)Levy, Seo, Choi, and Zettlemoyer]{levy2017zsre}
Levy, O., Seo, M., Choi, E., and Zettlemoyer, L.
\newblock Zero-shot relation extraction via reading comprehension.
\newblock \emph{arXiv preprint arXiv:1706.04115}, 2017.

\bibitem[Lewis et~al.(2019)Lewis, Liu, Goyal, Ghazvininejad, Mohamed, Levy,
  Stoyanov, and Zettlemoyer]{lewis2019bart}
Lewis, M., Liu, Y., Goyal, N., Ghazvininejad, M., Mohamed, A., Levy, O.,
  Stoyanov, V., and Zettlemoyer, L.
\newblock Bart: Denoising sequence-to-sequence pre-training for natural
  language generation, translation, and comprehension.
\newblock \emph{arXiv preprint arXiv:1910.13461}, 2019.

\bibitem[Lewis et~al.(2020)Lewis, Perez, Piktus, Petroni, Karpukhin, Goyal,
  K{\"u}ttler, Lewis, Yih, Rockt{\"a}schel, et~al.]{lewis2020rag}
Lewis, P., Perez, E., Piktus, A., Petroni, F., Karpukhin, V., Goyal, N.,
  K{\"u}ttler, H., Lewis, M., Yih, W.-t., Rockt{\"a}schel, T., et~al.
\newblock Retrieval-augmented generation for knowledge-intensive nlp tasks.
\newblock \emph{Advances in Neural Information Processing Systems},
  33:\penalty0 9459--9474, 2020.

\bibitem[Lewis et~al.(2021)Lewis, Wu, Liu, Minervini, K{\"u}ttler, Piktus,
  Stenetorp, and Riedel]{lewis2021paq}
Lewis, P., Wu, Y., Liu, L., Minervini, P., K{\"u}ttler, H., Piktus, A.,
  Stenetorp, P., and Riedel, S.
\newblock Paq: 65 million probably-asked questions and what you can do with
  them.
\newblock \emph{Transactions of the Association for Computational Linguistics},
  9:\penalty0 1098--1115, 2021.

\bibitem[Maillard et~al.(2021)Maillard, Karpukhin, Petroni, Yih, O{\u{g}}uz,
  Stoyanov, and Ghosh]{maillard2021multi}
Maillard, J., Karpukhin, V., Petroni, F., Yih, W.-t., O{\u{g}}uz, B., Stoyanov,
  V., and Ghosh, G.
\newblock Multi-task retrieval for knowledge-intensive tasks.
\newblock \emph{arXiv preprint arXiv:2101.00117}, 2021.

\bibitem[Ni et~al.(2021)Ni, Qu, Lu, Dai, {\'A}brego, Ma, Zhao, Luan, Hall,
  Chang, et~al.]{ni2021t5xRetrieval}
Ni, J., Qu, C., Lu, J., Dai, Z., {\'A}brego, G.~H., Ma, J., Zhao, V.~Y., Luan,
  Y., Hall, K.~B., Chang, M.-W., et~al.
\newblock Large dual encoders are generalizable retrievers.
\newblock \emph{arXiv preprint arXiv:2112.07899}, 2021.

\bibitem[Paranjape et~al.(2021)Paranjape, Khattab, Potts, Zaharia, and
  Manning]{paranjape2021hindsight}
Paranjape, A., Khattab, O., Potts, C., Zaharia, M., and Manning, C.~D.
\newblock Hindsight: Posterior-guided training of retrievers for improved
  open-ended generation.
\newblock \emph{arXiv preprint arXiv:2110.07752}, 2021.

\bibitem[Petroni et~al.(2021)Petroni, Piktus, Fan, Lewis, Yazdani, Cao, Thorne,
  Jernite, Karpukhin, Maillard, Plachouras, Rockt{\"{a}}schel, and
  Riedel]{kilt}
Petroni, F., Piktus, A., Fan, A., Lewis, P. S.~H., Yazdani, M., Cao, N.~D.,
  Thorne, J., Jernite, Y., Karpukhin, V., Maillard, J., Plachouras, V.,
  Rockt{\"{a}}schel, T., and Riedel, S.
\newblock {KILT:} a benchmark for knowledge intensive language tasks.
\newblock In \emph{Proceedings of the 2021 Conference of the North American
  Chapter of the Association for Computational Linguistics: Human Language
  Technologies, {NAACL-HLT} 2021, Online, June 6-11, 2021}, 2021.

\bibitem[Piktus et~al.(2021)Piktus, Petroni, Karpukhin, Okhonko, Broscheit,
  Izacard, Lewis, O{\u{g}}uz, Grave, Yih, et~al.]{piktus2021oyster}
Piktus, A., Petroni, F., Karpukhin, V., Okhonko, D., Broscheit, S., Izacard,
  G., Lewis, P., O{\u{g}}uz, B., Grave, E., Yih, W.-t., et~al.
\newblock The web is your oyster--knowledge-intensive nlp against a very large
  web corpus.
\newblock \emph{arXiv preprint arXiv:2112.09924}, 2021.

\bibitem[Raffel et~al.(2020)Raffel, Shazeer, Roberts, Lee, Narang, Matena,
  Zhou, Li, and Liu]{raffel2020exploring}
Raffel, C., Shazeer, N., Roberts, A., Lee, K., Narang, S., Matena, M., Zhou,
  Y., Li, W., and Liu, P.~J.
\newblock Exploring the limits of transfer learning with a unified text-to-text
  transformer.
\newblock \emph{Journal of Machine Learning Research}, 21:\penalty0 1--67,
  2020.

\bibitem[Roberts et~al.(2022)Roberts, Chung, Levskaya, Mishra, Bradbury, Andor,
  Narang, Lester, Gaffney, Mohiuddin, Hawthorne, Lewkowycz, Salcianu, van Zee,
  Austin, Goodman, Soares, Hu, Tsvyashchenko, Chowdhery, Bastings, Bulian,
  Garcia, Ni, Chen, Kenealy, Clark, Lee, Garrette, Lee-Thorp, Raffel, Shazeer,
  Ritter, Bosma, Passos, Maitin-Shepard, Fiedel, Omernick, Saeta, Sepassi,
  Spiridonov, Newlan, and Gesmundo]{roberts2022t5x}
Roberts, A., Chung, H.~W., Levskaya, A., Mishra, G., Bradbury, J., Andor, D.,
  Narang, S., Lester, B., Gaffney, C., Mohiuddin, A., Hawthorne, C., Lewkowycz,
  A., Salcianu, A., van Zee, M., Austin, J., Goodman, S., Soares, L.~B., Hu,
  H., Tsvyashchenko, S., Chowdhery, A., Bastings, J., Bulian, J., Garcia, X.,
  Ni, J., Chen, A., Kenealy, K., Clark, J.~H., Lee, S., Garrette, D.,
  Lee-Thorp, J., Raffel, C., Shazeer, N., Ritter, M., Bosma, M., Passos, A.,
  Maitin-Shepard, J., Fiedel, N., Omernick, M., Saeta, B., Sepassi, R.,
  Spiridonov, A., Newlan, J., and Gesmundo, A.
\newblock Scaling up models and data with $\texttt{t5x}$ and $\texttt{seqio}$.
\newblock \emph{arXiv preprint arXiv:2203.17189}, 2022.

\bibitem[Shazeer \& Stern(2018)Shazeer and Stern]{shazeer2018adafactor}
Shazeer, N. and Stern, M.
\newblock Adafactor: Adaptive learning rates with sublinear memory cost.
\newblock In \emph{International Conference on Machine Learning}, pp.\
  4596--4604. PMLR, 2018.

\bibitem[Singh et~al.(2021)Singh, Reddy, Hamilton, Dyer, and
  Yogatama]{singh2021emdr}
Singh, D., Reddy, S., Hamilton, W., Dyer, C., and Yogatama, D.
\newblock End-to-end training of multi-document reader and retriever for
  open-domain question answering.
\newblock \emph{Advances in Neural Information Processing Systems}, 34, 2021.

\bibitem[Thakur et~al.(2021)Thakur, Reimers, R{\"u}ckl{\'e}, Srivastava, and
  Gurevych]{thakur2021beir}
Thakur, N., Reimers, N., R{\"u}ckl{\'e}, A., Srivastava, A., and Gurevych, I.
\newblock Beir: A heterogenous benchmark for zero-shot evaluation of
  information retrieval models.
\newblock \emph{arXiv preprint arXiv:2104.08663}, 2021.

\bibitem[Thorne et~al.(2018)Thorne, Vlachos, Christodoulopoulos, and
  Mittal]{thorne2018fever}
Thorne, J., Vlachos, A., Christodoulopoulos, C., and Mittal, A.
\newblock Fever: a large-scale dataset for fact extraction and verification.
\newblock \emph{arXiv preprint arXiv:1803.05355}, 2018.

\bibitem[Voorhees(2001)]{voorhees2001philosophy}
Voorhees, E.~M.
\newblock The philosophy of information retrieval evaluation.
\newblock In \emph{Workshop of the cross-language evaluation forum for european
  languages}, pp.\  355--370. Springer, 2001.

\bibitem[Yang et~al.(2018)Yang, Qi, Zhang, Bengio, Cohen, Salakhutdinov, and
  Manning]{yang2018hotpotqa}
Yang, Z., Qi, P., Zhang, S., Bengio, Y., Cohen, W.~W., Salakhutdinov, R., and
  Manning, C.~D.
\newblock Hotpotqa: A dataset for diverse, explainable multi-hop question
  answering.
\newblock \emph{arXiv preprint arXiv:1809.09600}, 2018.

\bibitem[Zamani et~al.(2022)Zamani, Diaz, Dehghani, Metzler, and
  Bendersky]{zamani2022retrieval}
Zamani, H., Diaz, F., Dehghani, M., Metzler, D., and Bendersky, M.
\newblock Retrieval-enhanced machine learning.
\newblock \emph{arXiv preprint arXiv:2205.01230}, 2022.

\bibitem[Zobel(1998)]{zovel1998reliability}
Zobel, J.
\newblock How reliable are the results of large-scale information retrieval
  experiments?
\newblock In \emph{Proc. of SIGIR}, 1998.

\end{thebibliography}
\bibliographystyle{icml2022}



\end{document}